\documentclass[10pt,twocolumn,letterpaper]{article}

\usepackage[pagenumbers]{cvpr} % To force page numbers, e.g. for an arXiv version

\usepackage{cuted} % for teaser
\usepackage{ulem}

\usepackage{mathtools} % Added

\definecolor{cvprblue}{rgb}{0.21,0.49,0.74}
\usepackage[pagebackref,breaklinks,colorlinks,allcolors=cvprblue]{hyperref}

\def\paperID{22468} % *** Enter the Paper ID here
\def\confName{CVPR}
\def\confYear{2026}

\title{Neural Gabor Splatting: Enhanced Gaussian Splatting with Neural Gabor for High-frequency Surface Reconstruction}

\author{Haato Watanabe\\
The University of Tokyo\\
{\tt\small heart.watanabe.research@gmail.com}
\and
Nobuyuki Umetani\\
The University of Tokyo\\
{\tt\small n.umetani@gmail.com}
}

\begin{document}

\maketitle

% Teaser
\begin{strip}
  \centering
  \includegraphics[width=\textwidth]{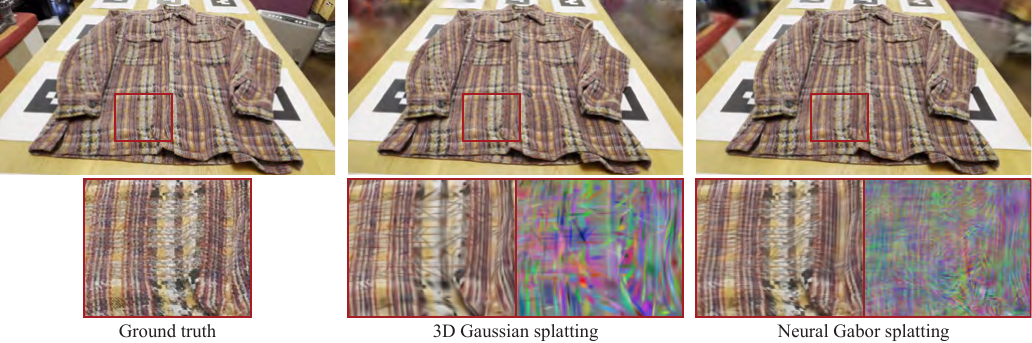}
  \captionof{figure}{
    While 3D Gaussian splatting struggles to represent fine texture using single-colored primitives—leading to numerous needle-like, slender elements—our neural Gabor splatting more effectively captures detail by encoding stripe patterns within each primitive. The bottom-right images show the primitive shapes with randomized colors. Both methods use an equal number of primitives.
    % Comparison on a \textsc{Shirt} from the High-Frequency Surface dataset. 
    % Our method reconstructs fine-scale patterns more faithfully than 3D Gaussian Splatting (3DGS) under the same primitive budget. 
    % Zoomed-in views highlight sharper texture details. 
    % Random-colored primitives visualize how our method encodes diverse color variations within each primitive, whereas 3DGS requires many primitives to approximate high-frequency appearance.
  }
  \label{fig:teaser}
\end{strip}

\begin{abstract}

Recent years have witnessed the rapid emergence of 3D Gaussian splatting (3DGS) as a powerful approach for 3D reconstruction and novel view synthesis. Its explicit representation with Gaussian primitives enables fast training, real-time rendering, and convenient post-processing such as editing and surface reconstruction. However, 3DGS suffers from a critical drawback: the number of primitives grows drastically for scenes with high-frequency appearance details, since each primitive can represent only a single color, requiring multiple primitives for every sharp color transition.
% ---
To overcome this limitation, we propose neural Gabor splatting, which augments each Gaussian primitive with a lightweight multi-layer perceptron that models a wide range of color variations within a single primitive. To further control primitive numbers, we introduce a frequency-aware densification strategy that selects mismatch primitives for pruning and cloning based on frequency energy.
% ---
Our method achieves accurate reconstruction of challenging high-frequency surfaces. We demonstrate its effectiveness through extensive experiments on both standard benchmarks, such as Mip-NeRF360 and High-Frequency datasets (e.g., checkered patterns), supported by comprehensive ablation studies.

%-------------------------------------------------------------------------

\end{abstract}
    
\section{Introduction}
\label{sec:introduction}

%-------------------------------------------------------------------------

% (First version)
% After 3D Gaussian splatting\cite{kerbl3Dgaussians} has merged, Gaussian splatting and the derived methods are widely used and well studied as novel view synthesis method using multiple photographs. Gaussian splatting is better than previous dominant method NeRF which exploit Multi layer perceptron to express a radians field and cumulative path tracing for rendering in rapidity of its training and rendering speed and editability owing to it's data format. Gaussian splatting express a scene with a set of Gaussian primitives which has a single color and reconstruct a whole scene optimizing all Gaussian's shape and color.

% (Second version)
Since the introduction of 3D Gaussian splatting (3DGS)~\cite{kerbl3Dgaussians}, Gaussian splatting and its variants have rapidly become popular methods for novel view synthesis from multi-view photographs. Compared to the neural radiance fields (NeRF)~\cite{mildenhall2020nerf}, which represent a radiance field with a multi-layer perceptron (MLP) and require costly volumetric ray marching, Gaussian splatting offers much faster training and rendering, as well as greater editability due to its explicit point-based representation.

%-------------------------------------------------------------------------

% (First version)
% Although Gaussian splatting superior in it's expression ability, it has deficiency in the size of data. Gaussian splatting uses many Gaussian primitives to reconstruct a whole scene and the number of these primitives and size of data are up to several millions and several Giga bytes. Many existing studies like ... worked this problem using data table, ... , etc to compress it's parameter expressions. However, Gaussian splatting tend to increase the size of data when it express high-frequency detailed scenes proved by Lu et al.\cite{lu2025poisonsplat}. This phenomenon is caused because each Gaussian primitives can express only one color. When high-frequency color is expressed, one or more Gaussians are needed for each color switching.

% (Second version)
Despite these advantages, Gaussian splatting suffers from a major drawback: its high data volume. A scene typically requires hundreds of thousands to millions of Gaussian primitives, often resulting in memory footprints of several gigabytes. Each primitive is parameterized by its position, scale, rotation, opacity, and color, and when scenes contain high-frequency details (e.g., sharp textures or checkerboard patterns), the number of required primitives increases sharply~\cite{lu2025poisonsplat}. This increase occurs because each Gaussian can encode only a single color for a given view direction, and multiple primitives are needed to represent frequent color transitions.

%-------------------------------------------------------------------------

% (First version)
% Several works tackled this problem. One promising direction is enhancing expression of each primitive. Watanabe et al.\cite{watanabe20253d} replaced Gaussian primitive with Gabor kernel to realize Gabor noise optimizaiton directly in 3D space. They embed several Gabor kernel into one primitive to represent complex pattern and achieved accurate fitting with less data even having more parameters. Chao et al. \cite{chao2025texturedgaussians} proposed Textured Gaussian to represent high-frequency details. Their method embed a small texture into each primitives and optimize it simultaneously with other parameters of the primitive. They archieved more accurate reconstruction of high-frequency detail part. Those works successfully enhanced expression ability for high-frequency detail embeding several color in each primitive. However those works are limited in expression ability because of their model formulation. For example, the expression ability of 3D Gabor splatting and Textured Gaussian are bounded by Gabor noise and hyperparameter texture size.

% compressionのサーベイを読んでcitationを入れる

% (Second version)
Several works have attempted to address this problem. Compression-based methods reduce storage by encoding parameters more efficiently~\cite{bagdasarian20253dgs}. Others aim to enhance the expressiveness of individual primitives. For example, Watanabe et al.~\cite{watanabe20253d} replaced Gaussians with 3D Gabor kernels, embedding multiple kernels per primitive to capture complex patterns. In contrast, Chao et al.~\cite{chao2025texturedgaussians} proposed textured Gaussians, attaching small learnable textures to each primitive to improve high-frequency fidelity. While effective, these approaches remain limited by their formulations: 3D Gabor splatting is constrained by the properties of Gabor noise, and textured Gaussians are bounded by the predetermined texture resolution.

%-------------------------------------------------------------------------

% (First version)
% We propose a method that enhances the expression ability of each primitive using Multi layer perceptron (MLP). Our method input local coordinate of each primitive to MLP and the MLP can express wide range of color pattern in one primitive. Because rendering local coordinate of each primitive consume large memory space, we also propose top-K contribution rendering method to supress memory footprint. 

% methodが固まり次第、内容を追加していく

% (Second version)
In this work, we introduce a new approach that improves the expressiveness of each primitive by embedding a lightweight MLP that maps local primitive coordinates and view-direction to color. This design enables a single primitive to represent a wide variety of patterns without being restricted to predefined bases or texture resolution. For efficient primitive number control, we further propose a frequency-aware densification strategy that utilizes a decomposed frequency-energy map to clone primitives within a specific frequency range. Consequently, our method achieves better reconstruction quality at high-frequency texture pattern areas with the same data budget, as shown in Fig.~\ref{fig:teaser}.

%-------------------------------------------------------------------------

% (First version)
% Our main contributions are as follws:
% \begin{itemize}
%  \item a simple approach to embed MLP to each primitive to enhance primitive's expression ability, 
%  \item a method to supress memory footprint in rendering that render only top-K contribution primitives, and
%  \item a comprehensive comparison in several dataset including our novel dataset that has high-frequency detailed scene.
% \end{itemize}

% (Second version)
Our contributions can be summarized as follows:
\begin{itemize}
 \item We present a simple yet effective approach to augment each Gaussian primitive with an MLP, significantly enhancing its expressiveness.
 \item We introduce a frequency-aware error-based densification strategy that reduces the number of primitives in optimization while preserving accuracy.
 \item We compare our method against existing approaches on public benchmarks (e.g., MipNeRF360) and high-frequency surface scenes, demonstrating superior reconstruction quality under constrained storage budgets.
\end{itemize}

\section{Related work}

%-------------------------------------------------------------------------

\subsection{Neural Representation for View Synthesis} 

Novel view synthesis (NVS) aims to generate images of a scene from unseen viewpoints given a set of input photographs. Early approaches, such as Structure from Motion (SfM)~\cite{brown2005unsupervised, snavely2006photo, agarwal2011building}, reconstruct sparse point clouds by matching image features to estimate camera poses. More recently, neural methods have dramatically improved reconstruction fidelity.
NeRF~\cite{mildenhall2020nerf, tancik2020fourier} models a continuous radiance field with an MLP and synthesizes images using ray marching, achieving impressive visual quality. NeRF leverages positional encoding to represent high-frequency details, but at the cost of expensive training and rendering.

To reduce computational cost, deferred neural rendering (i.e., neural textures)~\cite{thies2019deferred, meka2020_deep_relightable_textures, gao2020_deferred_neural_lighting} defines high-dimensional feature textures on geometry, where the features are decoded into RGB colors only for visible regions in screen space.
This design enables real-time rendering with greater expressive power.
Inspired by this approach, we equip each Gaussian primitive with a small MLP with its own weights.
The MLP outputs RGB colors from the input local primitive coordinates, allowing each primitive to represent diverse color patterns.

%-------------------------------------------------------------------------

\subsection{Gaussian Splatting and The Derived Methods}

3D Gaussian splatting (3DGS)~\cite{kerbl3Dgaussians, Yu2023MipSplatting} represents a scene with a set of Gaussian primitives and rasterizes them with alpha blending. Compared to NeRF, this explicit representation enables fast training and real-time rendering, and it naturally supports editing and downstream applications such as surface reconstruction \cite{huang20242d, guedon2023sugar}, dynamic scene modeling~\cite{wu20244d, yang2023gs4d, zhang2025mega}, sparse view reconstruction~\cite{li2025evpgs, chen2024mvsplat, charatan2024pixelsplat}, and scene editing~\cite{lee2025editsplat, pandey2025painting, gao2024real, watanabe2025SketchRodGS}. However, a typical scene may require hundreds of thousands to millions of primitives, leading to large memory footprints. To mitigate this, compression methods have been proposed~\cite{violante2025splat, wang2024contextgs, bagdasarian20253dgs}.

Another line of work improves the expressiveness of primitives themselves. For example, Huang et al.~\cite{huang20242d} proposed 2D Gaussian splatting, where Gaussian disks better approximate surfaces, and Li et al.~\cite{li2025hgs} introduced Half-Gaussian splatting to capture sharp edges. Despite these advances, data size remains a challenge due to the large number of parameters per primitive.
%-------------------------------------------------------------------------

\subsection{High-Frequency Detail for Gaussian Splatting} 

A key limitation of 3DGS is its inefficiency in modeling high-frequency details. Lu et al.~\cite{lu2025poisonsplat} observed that the number of primitives increases sharply for images with frequent color transitions, since each primitive can encode only a single color. Several approaches address this by allowing primitives to carry richer appearance information. 3D Gabor Splatting~\cite{watanabe20253d} and 3DGabSplat~\cite{zhou20253dgabsplat} replace Gaussians with augmented Gabor kernels, enabling direct optimization of high-frequency patterns in 3D. There is another way to capture fine-scale details that embed small textures into primitives~\cite{chao2025texturedgaussians, svitov2025billboard, rong2025gstex, weiss2024gaussian}, jointly optimized with other parameters. There is an approach to express sharp edges that tend to be erased because of the Gaussian's shape. Zhou et al.~\cite{zhou2025splat} proposed a method that expresses various types of primitive's shapes using a tiny MLP. Their method enables sharp edge switching in the reconstructed scene, while the primitive has only a single color. 
Recent works such as neural texture splatting~(NTS)~\cite{wang2025neural} and neural shell texture splatting~(NEST)~\cite{zhang2025neural} encode spatially varying color within each primitive, thereby enhancing expressiveness while reducing the number of primitives.

While existing detail-augmented primitive approaches enhance visual fidelity, they remain inherently limited by their underlying formulations:
Gabor splatting is constrained by predefined noise functions, texturing-based approaches are bounded by fixed texture resolution and directional dependence, and neural-texturing-based methods rely on fixed-capacity neural embeddings or predefined structures, which can limit flexibility under strict memory budgets.
In contrast, we embed an MLP conditioned on local coordinates, enabling each primitive to flexibly model a broader range of color variations without such restrictions.
Furthermore, we specifically target scenes with high-frequency (i.e., detailed) surfaces, where existing methods struggle to achieve sufficient reconstruction quality with a limited number of primitives.

% \input{sec/03_background}

%-------------------------------------------------------------------------

\begin{figure}[htbp!]
    \centering
    \includegraphics[width=\linewidth]{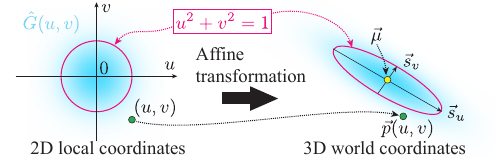}
    \caption{Coordinate transformation in 2D Gaussian splatting~\cite{huang20242d}}
    \label{fig:2dgs}
\end{figure}

\begin{figure*}[t!]
  \centering
    \includegraphics[width=\linewidth]{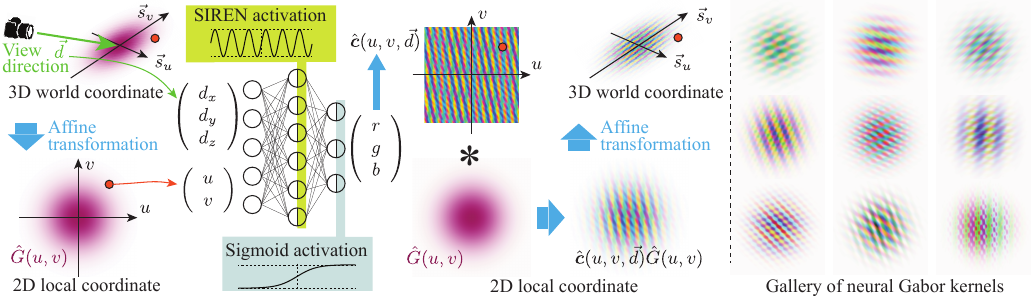}
  \caption{
    \textbf{Overview of the neural Gabor primitive.} 
    The output color is parameterized by local 2D coordinates obtained from the 2D Gaussian splatting primitive's affine transformation from 3D space.
    Given the 2D local coordinates and view direction, a lightweight MLP outputs RGB color through a SIREN activation function. 
    This design enables each primitive to model spatially varying and view-dependent appearance within a single primitive.
    % Workflow of our neural Gabor splatting.
  }
  \label{fig:neural-texture}
\end{figure*}

\section{Preliminaries: 2D Gaussian Splatting}

For self-consistency, this section briefly explains the existing 2D
Gaussian splatting~\cite{huang20242d} on which our algorithm is based.
The original Gaussian splatting~\cite{kerbl3Dgaussians} approximates the scene using a set of 3D Gaussian primitives, where a 3D Gaussian distribution of opacity represents each primitive. 
3D Gaussian distribution is formulated as
\begin{eqnarray}
  G(\vec{x}) &=& \exp\left(-\frac{1}{2} (\vec{x}-\vec{\mu})^T \Sigma^{-1} (\vec{x}-\vec{\mu})\right),\label{eq:gaussian}
  \label{eq:covariance}
\end{eqnarray}
where $\Sigma\in\mathbb{R}^{3\times 3}$ is the positive semi-definite covariance matrix. 
On the other hand, the 2D Gaussian splatting framework represents each primitive as a planar (flat) Gaussian distribution (i.e., $\Sigma$ is a rank-2 singular matrix). 
As illustrated in Fig.~\ref{fig:2dgs}, a 3D point $\vec{p}\in\mathbb{R}^3$ on the  plane is parameterized with the local 2D coordinate $(u,v)\in\mathbb{R}^2$ as
\begin{equation}
\vec{p}(u,v) = \vec{\mu} + u \vec{s}_u  + v\vec{s}_v,
\end{equation}
where the $\vec{\mu}\in\mathbb{R}^3$ is the center of the Gaussian primitive, $\vec{s}_u$ and $\vec{s}_v$ are the column vectors of $S\in\mathbb{R}^{3\times 2}$, where the covariance is decomposed as $\Sigma=S S^\top$.
The intersection between a camera ray and this plane is computed analytically, which enables precise and physically consistent evaluation of the Gaussian strength along the ray.
Given the intersection point's local coordinate $(u,v)$, the Gaussian kernel value is given as
\begin{equation}
G(u,v) = \exp \left\{-\frac{u^2+v^2}{2} \right\}.
\end{equation}

Each Gaussian primitive has spherical-harmonic coefficients $\theta_k$ for a view-dependent RGB expression; i.e., a Gaussian has a single color given a ray direction $\vec{d}\in\mathbb{R}^3$.

The color of each pixel is determined by alpha-blending, calculated by accumulating 2D Gaussian magnitude and colors from front to back. Gaussian strength is determined using local coordinates in each Gaussian as
\begin{eqnarray}
T_k &=& \prod_{j=1}^{k-1}\left\{
1 - \alpha_j G_j(u,v)\right\}, \\
\mathbf{c} &=& \sum_{k=1}^{K}\boldsymbol{c_k}(\theta_k, \vec{d})\alpha_k G_k(u,v)T_k \label{eq:alpha-blending},
\end{eqnarray}
where $T_k\in\mathbb{R}$ is the transmittance (influence of $k$-th kernel on the screen), $\alpha_k$ is the opacity of the primitive, and $K$ is the total number of the primitive.

\section{Method}

\subsection{Neural Gabor Splatting}

An overview of the neural Gabor primitive is illustrated in Figure~\ref{fig:neural-texture}. 
To represent complex color patterns and view-dependent appearance within a single primitive, we introduce a lightweight MLP that predicts RGB values from the primitive’s local coordinates and view direction. 
Each primitive has its own MLP parameters $\Theta_k$, enabling fine-grained modeling even with a compact memory capacity. We adopt a single-hidden-layer SIREN~\cite{sitzmann2020implicit} architecture with six hidden neurons, where the sinusoidal activation implicitly performs positional encoding and allows high-frequency signal representation without explicit encoding. 
Because the MLP jointly processes both spatial coordinates and view direction via nonlinear transformations, it can inherently learn view-dependent reflectance effects without requiring explicit spherical-harmonic (SH) bases.

Similar to 2DGS, the color of each pixel is computed as the weighted sum of analytically rasterized Gaussian primitives as
\begin{equation}
  \mathbf{c} = \sum_{k}^K{
        \hat{\boldsymbol{c}}_k(\Theta_k, u, v, \vec{d}) \alpha_k \hat{G}}_k
    T_k,
  \label{eq:alpha-blending-neural-texture}
\end{equation}
where $\hat{\mathbf{c}}_k\in\mathbb{R}^3$ denotes the color predicted by the neural Gabor, $\alpha_k$ the opacity, and $\hat{G}_k$ the normalized 2D Gaussian kernel.
The color prediction $\hat{\mathbf{c}}_k$ is obtained from a per-primitive MLP defined as
\begin{equation}
  \hat{\boldsymbol{c}}_k = \mathrm{Sigmoid}\left[
        \mathbf{\bar{W}}_k \sin\left\{
            \omega_0(\mathbf{W}_k \mathbf{y} + \boldsymbol{b}_k)        \right\}
            + \bar{\boldsymbol{b}}_k
    \right],
  \label{eq:siren}
\end{equation}
where $\mathbf{y} = (u, v, \vec{d})\in\mathbb{R}^5$ is the concatenation of the local coordinates and view direction, and $\omega_0\in\mathbb{R}$ is the frequency parameter, set to 30 in our implementation.
This formulation enables the network to capture nonlinear interactions between local surface coordinates and viewing directions, effectively learning complex specular and anisotropic effects that are difficult to express with fixed SH bases.

Unlike discrete texel-based textures, our neural Gabor splatting provides a continuous representation that is independent of texture resolution. 
Since each primitive maintains its own tiny MLP, the method achieves high representational power while significantly reducing the total number of primitives required. Consequently, the overall data footprint is compressed compared to using multiple single-color primitives in standard 2D or 3D Gaussian splatting.

\subsection{Frequency-aware Densification Strategy}

% 3DGSの高精度はGaussian primitiveを密にするdendification strategyに寄るところが大きい。しかしながら、従来のAdaptive Densification Methodはdensification対象のprimitiveの中心点位置のgradient値に依存しており、色の切り替わりが激しい提案手法のneural textured primitiveにそのまま適用した場合、必要以上にdensificationを行い、かえってデータ数を増加させてしまうことが判明した。そこでError based densification~\cite{}に着想を受けたFrequency Aware Densification Strategyを導入する。Error based densificationはレンダリング画像とGT画像間のエラーを3D空間中のprimitiveにinverse projectionすることで、densification対象のprimitiveを判定する手法である。カメラ \pi でレンダリングした時の各primitive \gamma_kのエラーのは以下のように定義される。per-pixel error \varepsilon_\pi(\boldsymbol{u})がprimitiveのcontrinution w_k^\pi(\boldsymbol{u})を用いて 各primitive \gamma_kに蓄積される。

The superior rendering quality of 3DGS primarily arises from its adaptive densification strategy, which allocates a higher density of Gaussian primitives to regions exhibiting large reconstruction errors.
However, the conventional gradient-based adaptive densification method depends on the gradient magnitude at the primitive center.
When directly applied to our neural Gabor primitives—whose colors may change abruptly due to learned high-frequency appearance—it tends to perform excessive densification, thereby increasing the number of primitives unnecessarily.  

\begin{figure*}[t!]
  \centering
   \includegraphics[width=\linewidth]{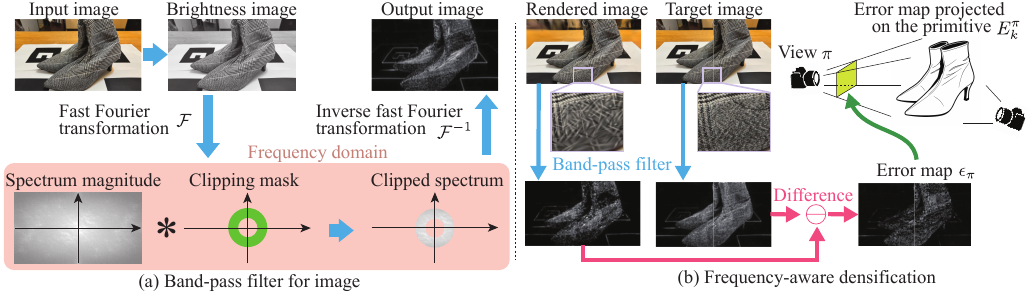}

   \caption{
        \textbf{Overview of the frequency-aware densification strategy.} 
        (a) Band-limited frequency components are extracted from both the rendered image and the ground truth using FFT, band-pass filtering, and inverse FFT, followed by local averaging to obtain a robust frequency-domain error map. 
        (b) The per-pixel frequency error is projected onto each Gaussian primitive based on its contribution, and primitives with high errors are selected for cloning or splitting. 
        This process allocates more primitives to regions with insufficient high-frequency representation while suppressing unnecessary growth.
        % Overview of our frequency-aware densification strategy.
   }
   \label{fig:frequency_map}
\end{figure*}

To address this issue, we introduce a \textit{frequency-aware densification strategy} (see~Fig.~\ref{fig:frequency_map}), inspired by the error-based densification method~\cite{rota2024revising}.  
Error-based densification determines which primitives should be cloned or split by inversely projecting the rendering error between the rendered image and the ground-truth image back into 3D space.  
For a given camera view $\pi$, the per-primitive error $E_k^\pi$ of each primitive $\gamma_k$ is computed by accumulating the per-pixel error $\varepsilon_\pi(\boldsymbol{u})$ weighted by its contribution $w_k^\pi(\boldsymbol{u})$, as 

\begin{equation}
  E_k^\pi \coloneqq 
    \sum_{
        \boldsymbol{u} \in \mathrm{Pixels}}{
        \varepsilon_\pi(\boldsymbol{u}) w_k^\pi(\boldsymbol{u}).
    }
  \label{eq:error-based-densification}
\end{equation}

%-------------------------------------------------------------------------

% エラーにはSSIMのような視覚指標の代わりにfrequency mapを用いた。レンダリング画像とGT画像から特定のバンド幅のfrequency mapを算出し、その差分を取ることで、周波数領域が足りていない空間部分を発見し、効率的に埋めることが可能となる。特定バンド幅のfrequency mapの算出にはFast Fourier TransformとInverse Fast Fourier Transformを使用し、frequency mapにaverage filterを掛けてから差分を取ることで微細な方向の違いにrobustなエラーを構築した。バンド幅は{(0.01, 0.10), (0.10, 0.20), (0.20, 0.40)}、average filerのkernel sizeは17を用いた。clone or splitするprimitiveのmlpのweight parameterは元のprimitiveのmlpのものをコピーし、opacity correctionは~\cite{}に従った。カメラ\piでレンダリングした画像I_\piに対するErrorの定義はGT画像GT_\piを用いて以下のように表される。Avgはlocal mean filtering、FFTとIFFTはFast Fourier TransformとInverse Fast Fourier Transformを意味する。

% Figure~\ref{fig:frequency-map}はEq.~\ref{eq:frequency-error}によって計算されたfrequency-domain error mapsを表している。smooth surfacesのような低周波領域はゼロに近い反応である一方、テクスチャの切り替わりやパターンの発生箇所のようなhigh-frequency領域は強い反応を示しており、重要な構造に反応する選択的densificationを可能にする。
% 実装上はfrequency-domain computationは計算オーバーヘッドが存在するものの、GPU上で並列して行われ、かつ100 iterationに1回のdensificationのみ計算されるので大きな負担にはならない。

Instead of using perceptual metrics such as SSIM, we define $\varepsilon_\pi(\boldsymbol{u})$ in the frequency domain.  
Specifically, frequency maps are computed from both the rendered image $I_\pi$ and the ground-truth image $GT_\pi$ within specified frequency bands, and their difference reveals regions lacking frequency components.  
This enables selective densification in areas with insufficient high-frequency content.  
We use Fast Fourier Transform (FFT) and its inverse (IFFT) to extract band-limited frequency components via a cropping operator, followed by a local mean filtering operator $\mathrm{Avg}$ to build a robust error metric less sensitive to directional misalignment.  
We use three frequency bands $\{(0.01, 0.10), (0.10, 0.20), (0.20, 0.40)\}$ and an averaging kernel size of $17$.  
For cloned or split primitives, the MLP weight parameters are copied from their parent, and opacity correction follows~\cite{rota2024revising}.  
The per-pixel frequency-domain error for a given camera $\pi$ is defined as

\begin{equation}
\begin{aligned}
  \varepsilon_\pi =
  \Bigl\|
    &\mathrm{Avg}\bigl(
      \mathcal{F}^{-1}(\mathrm{Bandcrop}(\mathcal{F}(GT_\pi)))
    \bigr) \\
    &-
    \mathrm{Avg}\bigl(
      \mathcal{F}^{-1}(\mathrm{Bandcrop}(\mathcal{F}(I_\pi)))
    \bigr)
  \Bigr\|_1,
  \label{eq:frequency-error}
\end{aligned}
\end{equation}
where $\mathcal{F}$ and $\mathcal{F}^{-1}$ denote the Fast Fourier Transform and its inverse, respectively, and $\mathrm{Avg}$ indicates local mean filtering.  
This formulation provides a frequency-sensitive error measure that promotes densification in under-represented frequency regions while suppressing redundant growth in well-reconstructed areas.  

%Figure~\ref{fig:frequency_map} visualizes the frequency-domain error maps computed by~\eqref{eq:frequency-error}.  
%Low-frequency regions, such as smooth surfaces, exhibit near-zero responses, whereas high-frequency regions—such as texture transitions or specular highlights—show strong activations, enabling selective densification focused on perceptually important structures.  

In practice, every 100 iterations, we randomly sample 20 training views and accumulate per-primitive errors across these views, which are jointly processed as a single GPU batch.
Densification candidates are determined using a threshold of 0.01.
The frequency-domain computation introduces only a minor overhead, as both the forward and inverse Fourier transforms ($\mathcal{F}$ and $\mathcal{F}^{-1}$) are highly parallelizable on modern GPUs.
Moreover, since densification is performed only once every 100 iterations, the overall additional cost remains negligible relative to the main training loop.

We also replaced the hard opacity reset with the gradual opacity reset proposed by Bulò et al.~\cite{rota2024revising}.
However, we did not adopt their alpha-compositing loss, as it noticeably degraded the visual quality by introducing undesirable artifacts.

%-------------------------------------------------------------------------

\subsection{Optimization}

% Original 3DGS~cite{}の手法に従って、$L_1$とSSIMをLossに用いて~\ref{eq:optimizaion}最適化を行った。

Following the original 3D Gaussian splatting (3DGS) method~\cite{kerbl3Dgaussians}, we optimized the model using a combination of the $L_1$ loss and SSIM loss as
\begin{equation}
\text{Loss} = \lambda L_1 + (1 - \lambda) L_{\text{SSIM}}.
\label{eq:optimizaion}
\end{equation}

% 我々の提案手法は単一のprimitiveで複数色のパターンを表現することであり、初期状態からパターンを表現する能力があるneural texture primitiveを配置し学習を行うことで、primitiveが小さくなり単色部分にfittingする減少を防止した。MLPの初期weightはSIREN model~cite{}の初期化手法に従い、$w_i \sim \mathcal{U}(-\sqrt{6 / n}, \sqrt{6 / n})$とした。

Our proposed method represents multiple color patterns within a single primitive.
By initializing and training neural texture primitives capable of expressing such patterns from the beginning, we prevent the primitives from collapsing into small, monochromatic regions during optimization.
The MLP weights were initialized following the SIREN model~\cite{sitzmann2020implicit} initialization scheme.

% 各GaussianはSH係数を持っており、視線角度により単色のRGBが評価される。言い換えると、１つの視線に対して単色のみを持つ。
% レンダリング時にはscreen space中における2D Gaussianの共分散行列がJacobianを用いて計算される。
% （Jacobianの数式）
% In rendering process, covariance matrix of 2D Gaussian in screen space is calculated using Jacobian which is formulated as \ref{eq:jacobian}.
% % （Jacobianの数式）
% \begin{equation}
%   \Sigma' = JW \Sigma W^TJ^T
%   \label{eq:jacobian}
% \end{equation}

% 各ピクセルの色は2D Gaussianのstrengthとcolorの積をカメラ方向にfront-to-back方向にalpha-blendingで蓄積することにより各ピクセルの色が決定される。Gaussianのｓtrengthはlocal-coordinateを用いて求められる。我々はこのlocal-coordinateをpositional encodingしてMLPに入力する。
% （alpha-blendingの数式）（local-coordinateを入れる）

% Backwardでは画像Lossを最小化する対する各パラメータのgradientが計算される。
% 我々はGaussian strengthの計算に用いられている各Gaussianのlocal coordinateとSH係数の代わりに追加したfeature vectorをMLPに入れることで１つのprimitive内で多様な色の変化を可能にする。（ここのロジックは変えるかも）
\begin{figure*}[t!]
  \centering
   \includegraphics[width=\linewidth]{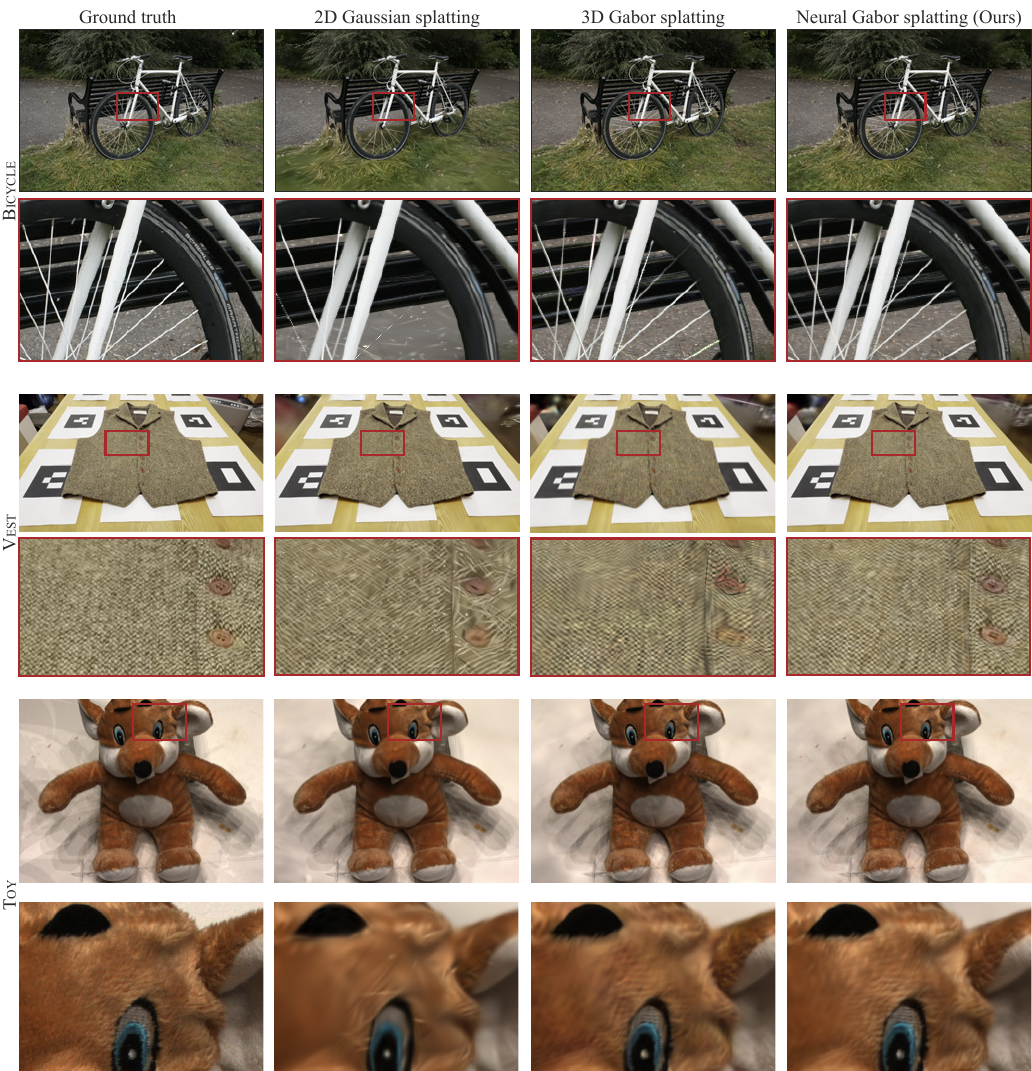}
   \caption{\textbf{Qualitative NVS results of benchmark and High-Frequency datasets.} Novel view rendering results of outside and inside scenes. Our method provides a clear, sharp visual of the fine details, despite the same amount of data.}
   \label{fig:visual_comparison}
\end{figure*}

\section{Result}

%-------------------------------------------------------------------------

\subsection{Implementation}

Based on the published code of 2DGS~\cite{huang20242d}, we implement our neural Gabor splatting~\footnote{ Repository for our published code:  \url{https://github.com/haato-w/neural-gabor-splatting}}.
We conducted all experiments on GeForce RTX 3090 GPU.
%
% The training takes roughly \textbf{~} min for 3DGS, \textbf{~} min for 2DGS and \textbf{~} min for our Neural Gabor splatting.
%
The runtime speed of our method is 30—500 frames per second (FPS) for all scenes.
%
% And the training speed is almost doubled of 2DGS. 
% The increase in training time can be attributed to the larger number of AtomicAdd operations required for weight updates.
% Further details, including the per-dataset averages, are provided in the supplemental material.
The training time is approximately twice that of 2DGS, mainly due to the increased number of atomicAdd operations required for weight updates.
% 
%We encourage the reader to see the animations in the supplemental video. 
% Please see the supplemental video for more detail.

%-------------------------------------------------------------------------

\subsection{Datasets}

% in-the-wildシーンの再構築精度を示すためのベンチマークデータセットとしてMip-NeRF360データセット~cite{}の全7シーンとDTUデータセットの4シーンを用いて評価を行った。加えて、high-frequencyなテスクチャを持つシーンに対する評価を行うためにhigh-frequency surface dataset~cite{}を用いた実験も行った。全てのデータセットを我々の環境で評価した。
% 各手法とデータセットにおいてprimitiveの初期個数と最大個数が設定した。最大個数は手法間でデータ量が同じになるように設定した。それぞれの手法とデータセットにおける設定値の詳細はsupplemental mateiralを参照してください。
% 我々の手法は少数のprimtiveでhigh-frequencyな見た目を持つシーンの再構築を高精度に行うことを目的としているため、少ないdensificationの回数で最大primitive数に到達し、一定上のiterationからは精度の向上率が大幅に低下する。そこで実験では全データセットにおいて20,000 iterationで行った。
% 3DGS, 2DGS, 3D Gabor splatting, our methodにおいてデータセットの画像は1600x1066pxにリサイズにしてトレーニングと評価を行った。、
% 初期点はデータセットのSfM point cloudからランダムサンプリングをした。3DGS, 2DGSのGaussianの中心位置のgradientを用いたdensification strategyがgradientが大きいものから優先的にdensification対象とし、上限以上のprimitiveは超えないようにした。pruning strategyは既存のものを採用し、primitive数が上限に近くなるとdeisificationとpruningが均衡するようになる。

To evaluate reconstruction accuracy on in-the-wild scenes, we used all seven scenes from the Mip-NeRF360 dataset~\cite{barron2022mip}, four scenes from the DTU dataset~\cite{jensen2014large} and two scenes from Tanks and Temples dataset~\cite{knapitsch2017tanks} as benchmark datasets.
In addition, to assess performance on scenes with high-frequency textures, we conducted experiments using the High-Frequency dataset~\cite{watanabe20253d}.
% All datasets were evaluated under our experimental environment.

%-------------------------------------------------------------------------

\subsection{Experimental Setup}

For each method and dataset, we set the initial and maximum numbers of primitives.
The maximum number was chosen to keep the total data size consistent across methods.
Detailed parameters for each dataset and method are provided in the supplemental material.

Since our method aims to accurately reconstruct scenes with high-frequency appearance using a small number of primitives, it reaches the maximum primitive count with relatively few densification steps.
Beyond a certain number of iterations, the accuracy improvement saturates.
Therefore, all experiments were performed with 20k training iterations.

For 3DGS, 2DGS, 3D Gabor splatting, and our method, input images were resized to 1600×1066 pixels for both training and evaluation.
The initial points were randomly sampled from the structure-from-motion (SfM) point cloud of each dataset.
In 3DGS and 2DGS, the densification strategy prioritized primitives with larger position gradients, while ensuring that the total number of primitives did not exceed the predefined limit.
The pruning strategy followed the original implementation, resulting in a balance between densification and pruning as the number of primitives approached the maximum threshold.

%-------------------------------------------------------------------------

\begin{table*}[t]
  \centering
  \caption{
      % \textbf{Quantitative comparisons of different novel view synthesis methods.} 
      % We emphasize the best-performing model in terms of different metrics (PSNR ↑ / SSIM ↑ / LPIPS ↓). We set the maximum primitive number of 3DGS and 2DGS equal to our method data size. Our model has better scores for almost all of the metrics, which means our method expresses scene representations efficiently. The bottom is a comparison between 3D Gabor splatting and our method ablates view-dependency. While 3D Gabor splatting can reconstruct scenes accurately comparable to our method, our method can integrate view-direction and local color pattern efficiently. 3DGS* and 2DGS* have a max primitive number limitation, 3D Gabor* is densified with our frequency-aware densification strategy, and Our* is omitted view-direction for fair comparison.
    \textbf{Quantitative comparison of novel view synthesis methods.}
    Best results are highlighted (PSNR ↑ / SSIM ↑ / LPIPS ↓).
    The maximum number of primitives for 3DGS and 2DGS is set to match the data size of our method.
    The bottom rows compare 3D Gabor Splatting and our method without view-dependent modeling.
    3DGS* and 2DGS* denote models with limited primitive counts, 3D Gabor* uses frequency-aware densification, and Ours* disables view-dependent components.
  }
  \label{tab:metrics}
  \resizebox{\textwidth}{!}{
  % \resizebox{0.85\textwidth}{!}{ % tableを小さくした
  \begin{tabular}{lcccc}
    \toprule
    Method & High-Frequency~\cite{watanabe20253d} & Mip-NeRF 360~\cite{barron2022mip} & DTU~\cite{jensen2014large} & Tanks and Temples~\cite{knapitsch2017tanks} \\
    \midrule
    3DGS* & 23.97 / 0.8335 / 0.2769 & \textbf{27.23} / 0.8005 / 0.2931 & 28.80 / 0.8467 / 0.4140 & 21.72 / 0.7673 / 0.3040 \\
    2DGS* & 23.91 / 0.8279 / 0.2855 & 26.47 / 0.7804 / 0.3197 & 28.98 / 0.8416 / 0.4171 & 21.12 / 0.7433 / 0.3375 \\
    % Chao et al. & --- & --- & --- \\
    \textbf{Ours} & \textbf{26.49} / \textbf{0.8808} / \textbf{0.2115} & 26.98 / \textbf{0.810} / \textbf{0.2521} & \textbf{29.15} / \textbf{0.8519} / \textbf{0.3705} & \textbf{21.93} / \textbf{0.7755} / \textbf{0.2762} \\
    \midrule
    3D Gabor* & 25.28 / 0.8530 / 0.2449 & \textbf{26.76} / \textbf{0.8265} / \textbf{0.2388} & 28.91 / \textbf{0.8573} / 0.3800 & 21.64 / 0.7757 / 0.2802 \\
    \textbf{Ours}* & \textbf{26.02} / \textbf{0.8776} / \textbf{0.2163} & 26.67 / 0.8172 / 0.2435 & \textbf{29.50} / 0.8570 / \textbf{0.3701} & \textbf{21.80} / \textbf{0.7811} / \textbf{0.2651} \\
    \bottomrule
  \end{tabular}
  }
\end{table*}

\subsection{Quantitative Results}

% table~\ref{tab:metrics}にPSNR、SSIM、LPIPSのメトリクススコアを示すHigh-frequencyデータセットは最大primitive数を20,000, 初期点群を5,000、DTUデータセットは最大primitive数を20,000, 初期点群を2,000、Mip-Nerf360は最大primitive数を200,000-3,000,000, 初期点群を4,000-600,000に設定した。3DGSと2DGSはデータ数が同じになるように最大primitive数を設定した。3D Gabor splattingは結果をfrequency-aware densificationを導入し、我々の提案手法と同じ最大primitive数に設定し、view-directionの入力を切り、隠れ層のニューロンを3つに変更した我々のモデルと比較した。

% 我々の提案手法は同じデータ量であるにも関わらず、3DGS, 2DGSと比較して全体的に全てのメトリクススコアで上回る結果で合った。これは提案手法が単一の色を持つprimitiveよりも幅広い表現力を持ち、色の切り替わりに対して効率的な表現ができていることをしてしている。

% table~\ref{tab:metrics}の下では3D Gabor splattingとの比較を行っている。SIRENモデルの表現力はGabor kernelを内包しているが、精度には大きな差は見られなった。これは我々の手法はneural networkを用いてpatern表現とview-dependency表現のintegrationを目指したため、完全にfairな比較はできない。

Table~\ref{tab:metrics} reports PSNR, SSIM, and LPIPS scores across all datasets.
For the High-Frequency dataset, we set the maximum number of primitives to 20,000 and initialized 5,000 points.
For the DTU dataset, we used a maximum of 20k primitives with 2k initial points.
For the Mip-NeRF360 dataset, the maximum primitive count ranged from 200k to 3,000k, depending on the scene, with 4k–600k initial points.
The maximum number of primitives for 3DGS and 2DGS was adjusted such that all methods used the same total data budget.
For 3D Gabor splatting, we introduced our frequency-aware densification strategy and matched its maximum primitive count to ours.
Additionally, to enable a reasonable comparison, we removed the view-direction input and reduced the hidden dimensionality to three neurons, aligning the model capacity more closely with our setting.

Despite using the same data budget, our method consistently outperforms both 3DGS and 2DGS across almost all of the evaluation metrics.
This improvement demonstrates that our neural texture primitives provide significantly higher expressive power than single-color primitives, enabling more efficient representation of high-frequency appearance variations.

The lower section of Table~\ref{tab:metrics} compares our approach with 3D Gabor Splatting.
Although a SIREN-based representation inherently subsumes Gabor kernels, the numerical differences remain modest.
This is expected, as our method integrates both pattern representation and view-dependent effects through a neural network, making a perfectly fair comparison difficult.

%-------------------------------------------------------------------------

\subsection{Qualitative Results}

% Figure~\ref{}にHigh-Frequency, Mip-NeRF360, DTU datasetにおけるNovel View Synthesisの結果のvisualを載せる。我々の手法は3DGS, 2DGSよりも同じデータ数にも関わらず、sharpかつclearなviewを生成できている。特にfurやfineなパターンの切り替わりを正確に捉えることができていることが分かる。加えて、反射のようなview-dependencyなものも扱うことができている。

Figure~\ref{fig:visual_comparison} presents qualitative novel-view synthesis results on the High-Frequency, Mip-NeRF360, and DTU datasets.
Despite using the same number of primitives, our method produces views that are noticeably sharper and clearer than those generated by 2DGS.
In particular, our approach faithfully captures fine-scale structures such as fur and rapid high-frequency texture transitions.
Moreover, the model effectively handles view-dependent effects, including subtle reflections, further demonstrating the expressive capacity of our neural texture primitives.
Please refer to the supplemental video for additional qualitative results.

%-------------------------------------------------------------------------

\subsection{Comparisons with Neural Splatting Methods}

We further compare our method against NTS~\cite{wang2025neural} and NEST~\cite{zhang2025neural} under similar data budgets across four models on the High-Frequency dataset and the \textsc{Bonsai} scene from Mip-NeRF360.
We follow the same settings as the budget-quality trade-off comparison in Sec.~\ref {sec:budget_quality_trade_off}.
%for High-Frequency scenes.
%
For Mip-NeRF360, we use the default NTS configuration and adjust the primitive count to match the total data size.
%
% Our primitive counts are fixed at 20k (High-Frequency) and 200k (Mip-NeRF360), and the reported scores are identical to those in the main paper.
Our primitive counts are fixed at 20k (High-Frequency) and 200k (Mip-NeRF360), consistent with the settings used in the main experiments.
%
% We observed similar trends in the DTU dataset. 
%
% We could not compare with 3DGabSplat due to the absence of a publicly available implementation.
%
As shown in Tab.~\ref {tab:init-max-num-others}, our method consistently outperforms NEST and NTS on High-Frequency dataset.
Specifically, we observed that NTS often fails to resolve detail texture (see supplementary document). 
%NTSはデータ数を少なくするとhigh-frequency部分でエラーっぽいものがでやすい"

%by preserving geometry while encoding high-frequency appearance within each primitive, whereas NEST and NTS suffer from reduced primitive capacity due to fixed-appearence overhead.

\begin{table}[b]
  \centering
  \caption{\textbf{Quantitative comparison under the same memory footprint.} Results are reported as PSNR / SSIM / LPIPS. High-Frequency scores are averaged over four scenes (we use \textsc{Bonsai} model for Mip-NeRF360*).}
  \label{tab:init-max-num-others}
  \begin{tabular}{@{}lcc@{}}
    \toprule
    Data             & High-Frequency & Mip-NeRF360* \\
    \midrule
    NEST        & 22.22 / 0.8588 / 0.2220 & 23.21 / 0.7889 / 0.3520 \\
    NTS         & 23.48 / 0.8139 / 0.3026 &  \textbf{29.49} / 0.9028 / 0.2544 \\
    \textbf{Ours}   & \textbf{26.49} / \textbf{0.8808} / \textbf{0.2115} & 29.18 / \textbf{0.9067} / \textbf{0.2412} \\
    \bottomrule
  \end{tabular}
\end{table}

% & \textbf{30.37} / \textbf{0.9227} / \textbf{0.2150} \\

%-------------------------------------------------------------------------

\subsection{Budget-quality Trade-off}
\label{sec:budget_quality_trade_off}
\begin{figure}[b!]
\centering
\includegraphics[width=\linewidth]{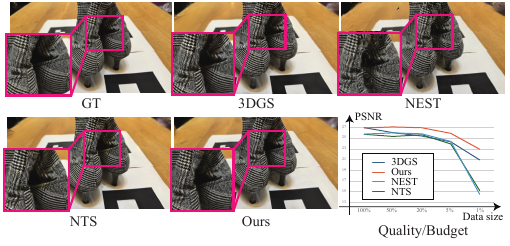}
\caption{Comparison under varying data budgets on the \textsc{Boots} model in High-Frequency dataset.}
\label{fig:rd-curve}
\end{figure}

We conduct a budget-quality study of the \textsc{Boots} model on the High-Frequency dataset by progressively reducing the data budget (100\% → 1\%), defined relative to the unconstrained 3DGS to show effectiveness of our method on strict memory budgets.
For each method, the maximum primitive count is adjusted to match the target footprint.
% ; for NEST and NTS, hash-grid and tri-plane resolutions are reduced accordingly.
For NEST and NTS, we reduce the capacity of their appearance representations to match the target memory footprint by lowering the resolution of the hash grid and tri-plane, respectively (e.g., NEST: 4 levels with a dictionary size of 16; NTS: tri-plane resolution of 96 with 8 channels).
As shown in Fig.~\ref{fig:rd-curve}, NEST and NTS degrade rapidly under tight budgets due to fixed appearance overhead, whereas our method is more robust in the low-budget regime and consistently outperforms 3DGS. 
We observe similar trends in SSIM.
Qualitatively, aggressive budget reduction causes artifacts or structural collapse in baselines, whereas our method preserves coherent geometry and appearance. Fig.~\ref{fig:rd-curve} also shows the reconstructed result on a 5\% data budget.

%-------------------------------------------------------------------------

\subsection{Training Time Analysis}
\label{sec:training-time}

% high-frequency surfaceのBootsデータセットでのtraining timeを載せる。budget間で比較。
% 100%, 50%, 20%はpruningの影響でデータ数が変わらないので20%以下で比較を行った。提案手法は3DGSと比較してMLP計算の分、レンダリング、バックワードともに遅くなるがそれ以外は変わらない。NESTと比較しても大きな乖離が無い計算時間である。

\begin{table}[h]
  \centering
  \caption{Training time (minutes) on the \textsc{Boots} model under different data budget.}
  \label{tab:training-time}
  \begin{tabular}{@{}lcccc@{}}
    \toprule
    Data Budget & 20\% & 5\% & 1\% \\
    \midrule
    3DGS   & 7.95 & 6.33 & 5.20 \\
    NEST   & 73.92 & 42.20 & 31.68  \\
    NTS   & 77.62 & 55.75 & 30.55 \\
    Ours   & 58.53 & 47.13 & 42.18 \\
    \bottomrule
  \end{tabular}
\end{table}

% We report training time on the high-frequency Boots dataset, focusing on low data budgets where primitive counts differ (see Tab.~\ref{tab:trainging-time}).
% While our method incurs additional cost from per-primitive MLPs compared to 3DGS, training time remains comparable in the low-budget regime due to reduced densification.
% The computational cost is also on par with NEST, indicating no excessive overhead.
% We report training time for the \textsc{Boots} model, focusing on low-data budgets in which the number of primitives differs across methods (Tab.~\ref{tab:training-time}).
%
% Compared with 3DGS, our method needs additional computational cost due to per-primitive MLP evaluation, resulting in longer training time.
% However, in the low-budget regime, the overall runtime remains practical because frequency-aware densification suppresses excessive growth of primitives.
% We also observe that the computational cost is comparable to NEST and NTS, indicating that our method does not introduce excessive overhead relative to existing neural splatting approaches. 
% We also provide comparisons of rendering speed under different data budgets in the supplemental material.
%
% We will include the detailed breakdown of the performance in the revision.

We report training time for the \textsc{Boots} model under low-data budgets where primitive counts differ across methods (Tab.~\ref{tab:training-time}).
Compared with 3DGS, our method incurs additional cost due to per-primitive MLP evaluations.
However, runtime remains practical in low-budget settings as frequency-aware densification suppresses excessive primitive growth.
The computational cost is comparable to NEST and NTS, indicating no significant overhead over prior neural splatting methods.
Rendering speed comparisons are provided in the supplemental material.

%-------------------------------------------------------------------------

\subsection{Ablation Study}

\begin{table}
  \centering
  \caption{\textbf{Ablation study of our frequency-aware densification strategy.} 
  While achieving nearly identical image quality to the error-based strategy, our frequency-aware variant offers improved controllability by selectively allocating densification to specific frequency bands.}
  \label{tab:ablation}
  \begin{tabular}{@{}lc@{}}
    \toprule
    Method & High-Frequency~\cite{watanabe20253d} \\
    \midrule
    Frequency-aware & 25.72 / \textbf{0.8619} / \textbf{0.2352} \\
    Error based & \textbf{25.95} / \textbf{0.8619} / 0.2376 \\
    Gradient based & 25.56 / 0.8534 / 0.2464 \\
    \bottomrule
  \end{tabular}
\end{table}

% 我々はError based densificaionに基づいて特定バンド領域のfrequencyをErrorに用いるFrequency-aware densification strategyを導入した。この章ではFrequency-aware densification strategyの特性を明らかにするためのablation studyを行う。
% 提案手法のNeural Gabor primitiveをFrequency-aware densification strategy, Error based densification strategy~\cite{} and 3DGS~\cite{}由来のprimitiveの中心座標のgradientによるdensification strategyを組み込み、High-Frequency datasetでシーンの学習を行った。table.~\ref{tab:ablation}はそれぞれの手法によるシーンのスコアを示す。Frequency-aware densification strategyのablation studyの結果を示す。
% Frequnecy-aware densification strategyはGradient based strategyと比較して高精度な再構築が可能であり、Error based densification strategyとほぼ同程度の精度である。しかしながら、提案手法は特定のバンド幅に所属するパターンに対してdensficationを掛けることができるため、densification可能なprimitiveの数が限られている場合、優先度の設定が可能となり再構築品質のコウンロールが容易であるという特徴がある。

We introduce a frequency-aware densification strategy derived from the error-based formulation by replacing the per-pixel error term with band-limited frequency errors.  
To clarify its characteristics, we integrate three densification strategies—our frequency-aware method, an error-based method~\cite{rota2024revising}, and a gradient-based method used in 3DGS~\cite{kerbl3Dgaussians}—into our neural Gabor primitive and train them on the High-Frequency dataset.

As shown in Table~\ref{tab:ablation}, our frequency-aware strategy reconstructs scenes with accuracy that is comparable to the error-based strategy and consistently outperforms the gradient-based strategy.  
More importantly, because it triggers densification only for primitives associated with specific frequency bands, it enables explicit prioritization when the number of densifiable primitives is limited.  
This property provides finer control over reconstruction quality, allowing the model to allocate capacity to visually critical high-frequency structures.

\section{Discussion}
% \textbf{Limitations and Future Work.} Our method assumes the presence of well-defined surfaces, such as clothing or walls, and therefore does not naturally extend to volumetric objects; extending the formulation to handle volume data remains an open problem. While effective for high-frequency content, the method may be inefficient for low-frequency scenes (i.e., scenes with small spatial color variation), in which MLP parameters can become redundant.
% %
% Adaptive control of MLP capacity could mitigate this issue.
% %
% In addition, the expressiveness of view-dependent appearance is ultimately limited by the capacity of the tiny per-primitive MLPs, suggesting that more sophisticated mechanisms may be required for materials with strong specular reflections.

% Another important direction is more aggressive scene compression. This includes merging primitives that share similar local appearance patterns, or introducing a codebook-style representation to quantize recurring primitive types. Finally, because our method captures rich local color patterns on surfaces, using these learned patterns for procedural material modeling represents an exciting avenue for future exploration.

\noindent\textbf{Limitations.}
While effective for high-frequency surface reconstruction, our method has several limitations.
It is not directly applicable to volumetric phenomena and extending it to dynamic scenes remains non-trivial.
Although our approach reduces primitive counts, it introduces additional computational overhead due to per-primitive MLP evaluation.
Moreover, for low-frequency scenes, the expressive capacity of the MLP may be underutilized.
Finally, the representation is still limited by the capacity of lightweight MLPs, and further improvements such as parameter sharing or codebook-based compression are promising directions.

% ----------

% primitive毎に個別のMLPをもたせ、primitiveのローカル座標とview-direciotnから色を推論することで単一のprimitive上で複雑な色パターンを表現可能であるNeural Gabor splttingを提案した。Frequecy-aware densification methodも導入し、少ないprimitive数でhigh-frequnecy surface objectを高品質に再構築することを可能にした。ベンチマークデータセットとhigh-frequnecy surfaceを含むHigh-frequnecy Surface datasetを用いて3DGS, 2DGS, 3D Gabor splattingと比較を行い、同じデータ量で高い再構築精度を実現することを示した。

\textbf{Conclusion.} We introduced neural Gabor splatting, a new representation that equips each Gaussian primitive with a lightweight MLP, enabling a single primitive to capture complex local color patterns depending on the view direction. We further incorporated a frequency-aware densification strategy, which selectively allocates primitives based on frequency-band errors, allowing high-frequency surface regions to be reconstructed with far fewer primitives.

Evaluations on standard benchmarks and the High-Frequency dataset show that our method outperforms 3DGS, 2DGS, and 3D Gabor splatting under comparable data budgets, achieving higher accuracy with far fewer primitives.

\section*{Acknowledgements}
We thank the anonymous reviewers for their valuable feedback. 
We also thank Chang Luo for providing access to his private GPU when conducting our experiments. 

{
    \small
    \bibliographystyle{ieeenat_fullname}
    \bibliography{reference}
}


% \newpage
% \input{supplemental_material}

\clearpage
\appendix
\section*{Supplementary Material}
% ---------------------------------------------------------
% CVPR Supplementary Material Template
% ---------------------------------------------------------
% \documentclass[10pt,twocolumn,letterpaper]{article}

% \usepackage{cvpr}              % CVPR style
% \usepackage{times}             % Times New Roman
% \usepackage{epsfig}
% \usepackage{graphicx}
% \usepackage{amsmath,amssymb}
% \usepackage{multirow}
% \usepackage{booktabs}
% \usepackage{enumitem}
% \usepackage{caption}
% \usepackage{subcaption}
% \usepackage{hyperref}
% \usepackage{xr}
% \externaldocument{main} % main.tex の名前（.auxを参照）

% Remove page numbers
% \pagestyle{empty}
% \thispagestyle{empty}

% ---------------------------------------------------------
% \begin{document}

% ---------------------------------------------------------
% \title{\vspace{-1.0cm}
% Neural Gabor Splatting: Enhanced Gaussian Splatting with Neural Gabor for High-frequency Surface Reconstruction\\[6mm]
% \textbf{Supplementary Material}
% }

% For review period:
% \author{Anonymous CVPR Submission\\
% Paper ID: 1234}

% For camera-ready:
% \author{Paper ID 22468}

% \maketitle
% \thispagestyle{empty}

% ---------------------------------------------------------
\section{Experiment Details}

To ensure a fair comparison across different methods, we constrained the number of primitives such that the total data size remained roughly the same. 
To make the data size similar, we defined the initial number of primitives and the maximum allowable number of them for each dataset and method.
Tables~\ref{tab:init-max-num-mipnerf} and \ref{tab:init-max-num-others} summarize these settings. 
Across all datasets, the number of primitives after training stays within 100–200 of the designated maximum. This small discrepancy arises from the interplay between densification and pruning.

\paragraph{Densification and Pruning.}
Densification was performed every 100 iterations.  
For 3DGS and 2DGS, we adopt the default densification thresholds from the original implementations.  
Densification continued until 15,000 iterations, after which further accuracy gains saturated and the primitive count had nearly reached the predefined maximum.
Once the predefined maximum was reached, new primitives were added only from those having the largest position gradients, ensuring that the primitive number does not exceed the limit.  
The pruning follows the default strategy.

For 3D Gabor splatting, achieving perfect fairness in data size is difficult because the method does not model view-dependent appearance.  
Therefore, instead of matching the total data size, we matched the \emph{number of primitives} to our method to ensure a balanced and interpretable comparison.  
Densification and pruning followed the same scheduling as our method.

\paragraph{Frequency-Aware Densification.}
For our method, frequency-aware densification was applied for 19,000 out of the 20,000 training iterations.  
During each densification step, we randomly sampled 20 training views, processed them as a batch, and applied FFT and inverse FFT filtering.  
The per-pixel frequency-domain error from each view was projected onto the corresponding primitives, and for each primitive, the maximum error over the 20 views was used as the densification score.
% %
% \textbf{We will publish our code and dataset upon acceptance.}

\paragraph{Datasets.}
For the DTU dataset, we used scenes \texttt{scan83}, \texttt{scan105}, \texttt{scan106}, and \texttt{scan114}.  
For Tanks and Temples, we evaluated the \texttt{train} and \texttt{truck} scenes.  
Across all datasets, we used a 1/8 split for testing, with the remaining images used for training.

\begin{table*}
  \centering
  \caption{\textbf{Initial and maximum primitive counts for each Mip-NeRF360 scene.}
  Left value indicates the initial primitive count; right value indicates the maximum count.}
  \label{tab:init-max-num-mipnerf}
  % \resizebox{\textwidth}{!}{
  \begin{tabular}{@{}lcccc@{}}
    \toprule
    Mip-NeRF 360 & 3DGS & 2DGS & \textbf{Ours} & 3D Gabor \\
    \midrule
    Bicycle, Garden, Stump    & 740k-3,700k & 740k-3,700k & 600k-3,000k & 600k-3,000k \\
    Bonsai, Counter, Kitchen, Room & 50k-243k & 50k-243k & 40k-200k & 40k-200k \\
    \bottomrule
  \end{tabular}
  % }
\end{table*}

\begin{table*}
  \centering
  \caption{\textbf{Initial and maximum primitive counts for High-Frequency~\cite{watanabe20253d}, DTU~\cite{jensen2014large}, and Tanks and Temples (tandt)~\cite{knapitsch2017tanks}.}
  Left value indicates the initial primitive count; right value indicates the maximum count.}
  \label{tab:init-max-num-others}
  \begin{tabular}{@{}lcccc@{}}
    \toprule
    Dataset         & 3DGS & 2DGS & \textbf{Ours} & 3D Gabor \\
    \midrule
    High-Frequency  & 6k-24k & 6k-24k & 5k-20k & 5k-20k \\
    DTU             & 2.5k-24k & 2.5k-24k & 2k-20k & 2k-20k \\
    train (tandt)   & 60k-356.3k & 60k-356.3k & 60k-300k & 60k-300k \\
    truck (tandt)   & 20k-118.8k & 20k-118.8k & 20k-100k & 20k-100k \\
    \bottomrule
  \end{tabular}
\end{table*}

% ---------------------------------------------------------

\section{Additional Quantitative Results}

% 各primitiveに個別のテクスチャを貼ることで高精細な再構築を可能にするTextured Gaussians~\cite{chao2025texturedgaussians}と我々の提案手法のQuantitative比較実験を行った。

% 我々の提案手法はTextured Gaussiansに全指標でスコアが上回った。
% Texture Gaussiansは通常のGaussian splattingでpre-trainingした後でtextureを貼ったGaussianをトレーニングすることで高精細な表現を出すことができる。しかし、十分なprimitiveが空間中にない場合、pre-trainingの段階でhigh-frequencyな色に過剰にfittingした結果細く潰れたprimitiveが大量に構成されてしまい、textureを貼っても本来のtextureの表現力を活かすことができない。我々の手法は多彩な色パターンを表現できるNeural Gabor primitiveをend-2-endで学習するため、high-frequencyな色部分に対して単色でfittingすることを避けることができるため少データでも高品質に再構築が可能である。

% 我々の提案手法のスコアはmain paperの再掲である。

% Textured GaussiansのテクスチャはRGBAを持ち、データ数を同じにするためにprimitive数は10,200、テクスチャ resolutionは4に設定した。

We further compare our method with Textured Gaussians~\cite{chao2025texturedgaussians}, which achieves high-fidelity reconstruction by assigning an individual texture map to each primitive. 
Table~\ref{tab:textured-gaussians-comparison} reports the quantitative results on the High-Frequency dataset~\cite{watanabe20253d}.

Across all metrics, our method outperforms textured Gaussians under the limited primitive budget. 
Although textured Gaussians improves appearance quality by training textured primitives after a pre-trained Gaussian splatting stage, this two-stage pipeline implicitly requires a sufficient number of primitives to cover fine-scale structures. 
When the primitive count is constrained, its pre-training stage tends to overfit high-frequency regions with overspecialized and elongated primitives, reducing the effective representational capacity of the subsequent texture optimization.

\begin{figure}
\centering
\includegraphics[width=\linewidth]{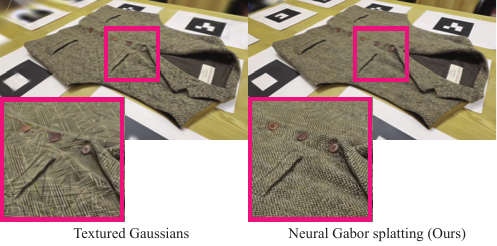}
\caption{\textbf{Visual comparison between Textured Gaussians and our method on High-Frequency dataset.} Our method expresses high-frequency texture successfully, while Textured Gaussians produces elongated monotonous color primitives due to the separation of initial primitive shape training and subsequent texture training.}
\label{fig:compare-textured-gaussians}
\end{figure}

In contrast, our method trains the neural Gabor primitives in an end-to-end manner, enabling each primitive to represent diverse color patterns and avoiding degenerate fitting behaviors in high-frequency regions (Fig.~\ref{fig:compare-textured-gaussians}). 
As a result, our model preserves complex appearance details even with substantially fewer primitives.

To ensure a fair comparison in terms of data size, we configured Textured Gaussians to use RGBA textures with a resolution of $4\times4$, and we set the number of primitives to 10,200. This setting matches the total data footprint of our method while respecting the constraints of the original Textured Gaussians formulation.

The quantitative results for our method in Table~\ref{tab:textured-gaussians-comparison} match those presented in the main paper.

\begin{table}
  \centering
  \caption{\textbf{Additional quantitative comparison between Textured Gaussians~\cite{chao2025texturedgaussians} and our method on the High-Frequency dataset.}
  Our method achieves consistently higher performance under the constrained primitive setting.}
  \label{tab:textured-gaussians-comparison}
  % \resizebox{\textwidth}{!}{
  \begin{tabular}{@{}lccc@{}}
    \toprule
     & PSNR ↑ & SSIM ↑ & LPIPS ↓  \\
    \midrule
    Textured Gaussians  & 23.57 & 0.7912 & 0.3203 \\
    \textbf{Ours}       & \textbf{26.49} & \textbf{0.8808} & \textbf{0.2115}  \\
    \bottomrule
  \end{tabular}
  % }
\end{table}

% ---------------------------------------------------------
\section{Additional Qualitative Results}

% Fig.~\ref{fig:add-qualitative-mipnerf} から
% in-the-wildシーンでも我々の手法は綺麗に再構築できる
% 文字のような細い線の変化も捉えることができる
% stumpのデータセットでは3DGSと同じデータ量であるにも関わらずよりGround Truth画像に近い見た目を再現できている
% 同じデータ量でも効率良くhigh-frequencyを表現できることが分かる。
% 3D Gabor splattingはhigh-ferequencyを表現できるものの、view-dependencyを考慮できないことからartifactが見て取れる。

% Fig.~\ref{fig:add-qualitative-others} から
% 服の模様のような複雑なパターンも効率良く表現できることがわかる
% ブーツのデータセットでは極めて細かい模様があるものの、我々の手法はblurされずに表現することができている。

We present additional qualitative novel-view synthesis results on the High-Frequency dataset, the Mip-NeRF 360 dataset, and the DTU dataset. We compare our method with 2D Gaussian splatting (2DGS) and 3D Gabor splatting. For fairness, we augment 3D Gabor splatting with our frequency-aware densification strategy.
\textbf{For all methods, the maximum number of primitives is constrained to match the same overall data size.}

\begin{figure*}
\centering
\includegraphics[width=\linewidth]{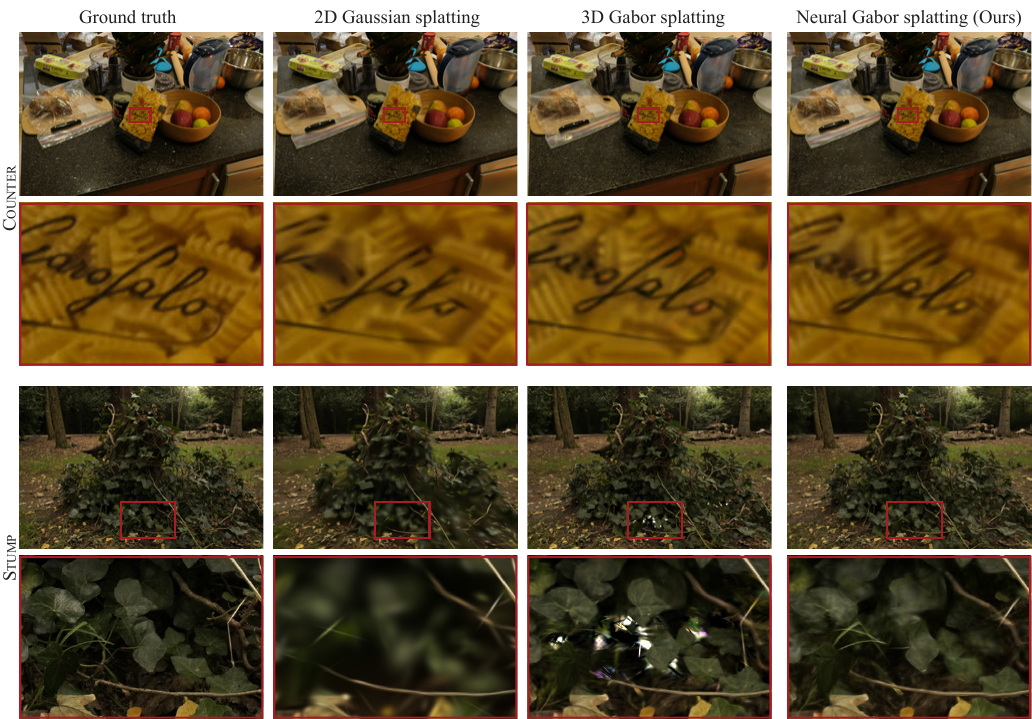}
\caption{\textbf{Additional qualitative comparisons on Mip-NeRF 360 with identical data size.} Our method produces higher visual fidelity compared to 2DGS and 3D Gabor Splatting. 3D Gabor Splatting tends to generate color artifacts (circled) due to the absence of view-dependent modeling.}
\label{fig:add-qualitative-mipnerf}
\end{figure*}

As shown in Fig.~\ref{fig:add-qualitative-mipnerf}, our method produces sharper and more faithful reconstructions than both 2DGS and 3D Gabor Splatting on the Mip-NeRF 360 dataset. In particular, 3D Gabor Splatting—while capable of representing high-frequency components—exhibits noticeable color artifacts due to its lack of view-dependent modeling.

\begin{figure*}
\centering
\includegraphics[width=\linewidth]{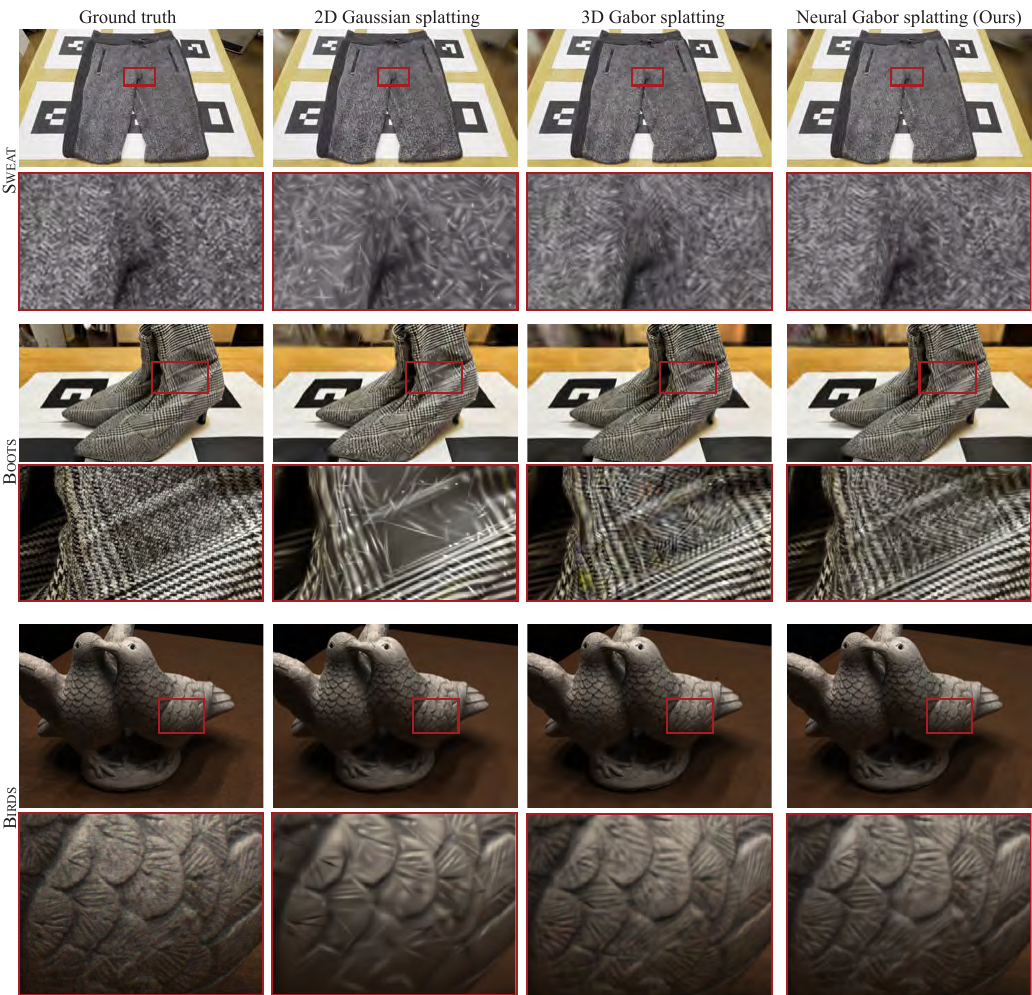}
\caption{\textbf{Additional qualitative comparisons on the High-Frequency and DTU datasets under identical data size.} Our method better preserves fine-grained patterns—such as intricate clothing textures and thin structures—especially when the primitive count is limited.}
\label{fig:add-qualitative-others}
\end{figure*}

Additional comparisons on the High-Frequency and DTU datasets are presented in Fig.~\ref{fig:add-qualitative-others}. Even under strict limitations on the number of primitives, our method successfully preserves fine-grained patterns—such as intricate clothing textures and thin structures—while competing methods suffer from blurring or loss of detail.

% 加えて、additional novel-view synthesis videoから
% 我々の手法は3次元空間をview-dependencyを含めて崩壊無く再構築できていることが分かる

Moreover, the supplemental novel-view synthesis videos further demonstrate that our method reconstructs 3D scenes without structural collapse even under view-dependent appearance changes, highlighting the stability of our representation across continuous camera movements.

% ---------------------------------------------------------
\section{Visual Comparisons with Neural Splatting Methods}

\begin{figure*}
\centering
\includegraphics[width=\linewidth]{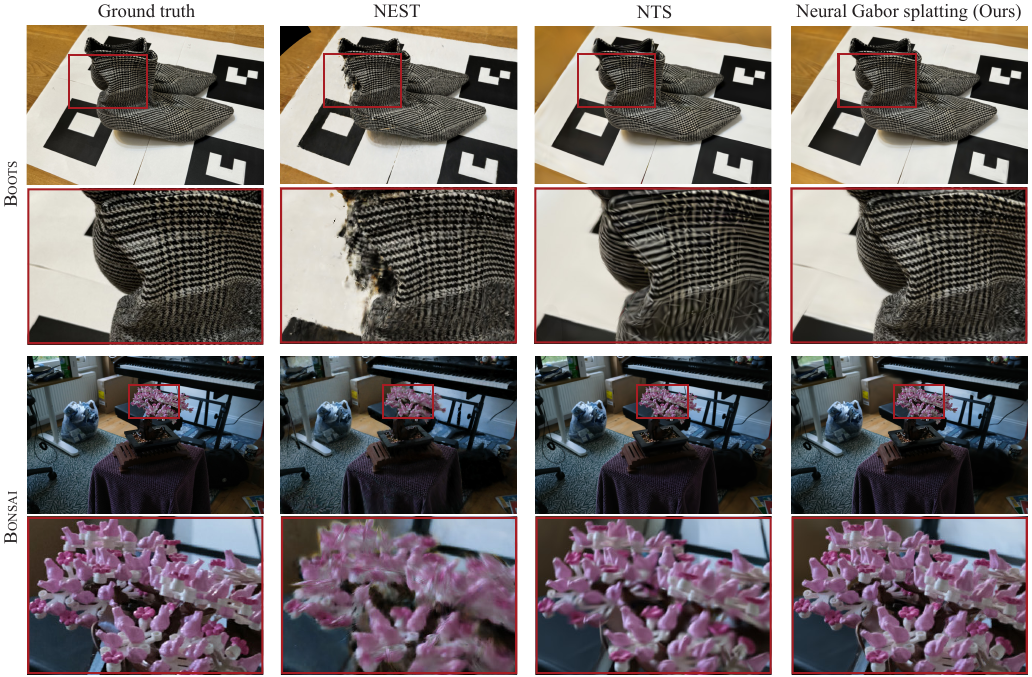}
\caption{
    % \textbf{Additional qualitative comparisons on the High-Frequency and DTU datasets under identical data size.} Our method better preserves fine-grained patterns—such as intricate clothing textures and thin structures—especially when the primitive count is limited.
    \textbf{Visual comparisons with NEST and NTS under matched memory budgets.}
    Results on the \textsc{Boots} scene from the High-Frequency dataset and the \textsc{Bonsai} scene from Mip-NeRF360 are shown.
    Each group includes the full image and a corresponding zoomed-in image highlighting fine-scale structures.
}
\label{fig:comparison-neural-texture}
\end{figure*}

We provide additional visual comparisons with NEST~\cite{zhang2025neural} and NTS~\cite{wang2025neural} under the same experimental settings as Sec.~5.6.
Fig.~\ref{fig:comparison-neural-texture} shows results on the \textsc{Boots} scene from the High-Frequency dataset and the \textsc{bonsai} scene from Mip-NeRF360, each with corresponding zoomed-in views.

We observe that NEST tends to exhibit noticeable visual degradation under tight memory budgets, often leading to structural inconsistencies or artifacts.
NTS is affected by suboptimal primitive configurations such as skinny Gaussians, resulting in insufficient fitting that cannot be fully compensated for by its neural texture representation.

In contrast, our method maintains more stable geometry and appearance across both scenes, with fewer visible artifacts under the same budget constraints.

% ---------------------------------------------------------
\section{Rendering Time}

We report rendering speed (FPS) under the same experimental setup as Sec.~5.8 of main paper (Tab.~\ref{tab:rendering-time}).
Due to per-primitive MLP evaluation, our method is slower than 3DGS.
However, it achieves significantly higher rendering speed than NEST and NTS, which rely on additional tri-plane or hash-grid computations.
This indicates that our method introduces moderate overhead while remaining efficient among neural splatting approaches.

\begin{table}[h]
  \centering
  \caption{Rendering time (FPS) on the \textsc{Boots} model under different data budget.}
  \label{tab:rendering-time}
  \begin{tabular}{@{}lcccc@{}}
    \toprule
    Data Budget & 20\% & 5\% & 1\% \\
    \midrule
    3DGS   & 242.8 & 349.7 & 467.3 \\
    NEST   & 12.25 & 24.55 & 43.86 \\
    NTS   & 11.69 & 15.01 & 51.89 \\
    Ours   & 85.32 & 116.8 & 145.4 \\
    \bottomrule
  \end{tabular}
\end{table}

% % ---------------------------------------------------------
% \section{Additional Ablation Studies}
% \subsection{Effect of Network Architecture}
% Describe your further analysis here. Provide extra tables or figures.

% \begin{table}[h]
% \centering
% \begin{tabular}{lccc}
% \toprule
% Method & PSNR ↑ & SSIM ↑ & LPIPS ↓ \\
% \midrule
% Ours (full) & 27.12 & 0.865 & 0.240 \\
% w/o module A & 26.48 & 0.852 & 0.265 \\
% w/o module B & 25.90 & 0.844 & 0.284 \\
% \bottomrule
% \end{tabular}
% \caption{Additional ablation results.}
% \end{table}

% {
%     \small
%     \bibliographystyle{ieeenat_fullname}
%     \bibliography{reference}
% }

% \end{document}

\end{document}